\pdfoutput=1

\documentclass[11pt]{article}

\newcommand{\veryshortarrow}[1][3pt]{\mathrel{%
   \hbox{\rule[\dimexpr\fontdimen22\textfont2-.2pt\relax]{#1}{.4pt}}%
   \mkern-4mu\hbox{\usefont{U}{lasy}{m}{n}\symbol{41}}}}

 \usepackage{hyperref}
\usepackage{EMNLP2023}
\usepackage{amsmath}
\usepackage{stfloats}
\usepackage{times}
\usepackage{latexsym}
\usepackage{graphicx}
\usepackage{makecell}
\usepackage{bm}
\usepackage{multirow}
\usepackage{booktabs}
\usepackage[T1]{fontenc}
\usepackage{arydshln}
\usepackage{amsfonts,amssymb}

\usepackage[utf8]{inputenc}

\usepackage{microtype}

\usepackage{inconsolata}

\newcommand\our{\textsc{VioLA}}

\title{\our{}: Unified Codec Language Models for Speech \\ Recognition, Synthesis, and Translation}

\author{Tianrui Wang$^*$, \  Long Zhou\thanks{\ \ Equal Contribution. Working in progress.} , \ Ziqiang Zhang, \ Yu Wu, \ Shujie Liu, \\ \textbf{Yashesh Gaur, \ Zhuo Chen, \ Jinyu Li,  \ Furu Wei}\\
Microsoft \\
\textit{ \{v-tianrwang, lozhou, v-ziqzhang, yuwu1, shujliu, yagaur, zhuc,jinyli, fuwei\}@microsoft.com } \\
} 

\begin{document}
\maketitle
\begin{abstract}

Recent research shows a big convergence in model architecture, training objectives, and inference methods across various tasks for different modalities.
In this paper, we propose \textbf{\our{}}, a single auto-regressive Transformer decoder-only network that unifies various cross-modal tasks involving speech and text, such as speech-to-text, text-to-text, text-to-speech, and speech-to-speech tasks, as a conditional codec language model task via multi-task learning framework.
To accomplish this, we first convert all the speech utterances to discrete tokens (similar to the textual data) using an offline neural codec encoder.
In such a way, all these tasks are converted to token-based sequence conversion problems, which can be naturally handled with one conditional language model. 
We further integrate task IDs (TID) and language IDs (LID) into the proposed model to enhance the modeling capability of handling different languages and tasks.
Experimental results demonstrate that the proposed \our{} model can support both single-modal and cross-modal tasks well, and the decoder-only model achieves a comparable and even better performance than the strong baselines.

\end{abstract}

\section{Introduction}

Recent years have revealed a significant convergence in model architectures, training techniques, and inference methods across research domains \cite{vaswani2017attention,devlin2018bert,brown2020language,wang2022image,gpt4}. 
Although different model frameworks, e.g., encoder-only BERT \cite{devlin2018bert}, decoder-only GPT \cite{radford2018improving}, and encoder-decoder based BART \cite{lewis2019bart} and T5 \cite{raffel2020exploring}, have made remarkable improvements in natural language processing (NLP) tasks, emerging advances like ChatGPT \cite{chatgpt} demonstrate that generative models have strong modeling abilities and potential to address various NLP tasks, especially for few-shot or zero-shot problems.

\begin{figure*}[t]
  \centering
  \includegraphics[width=15cm]{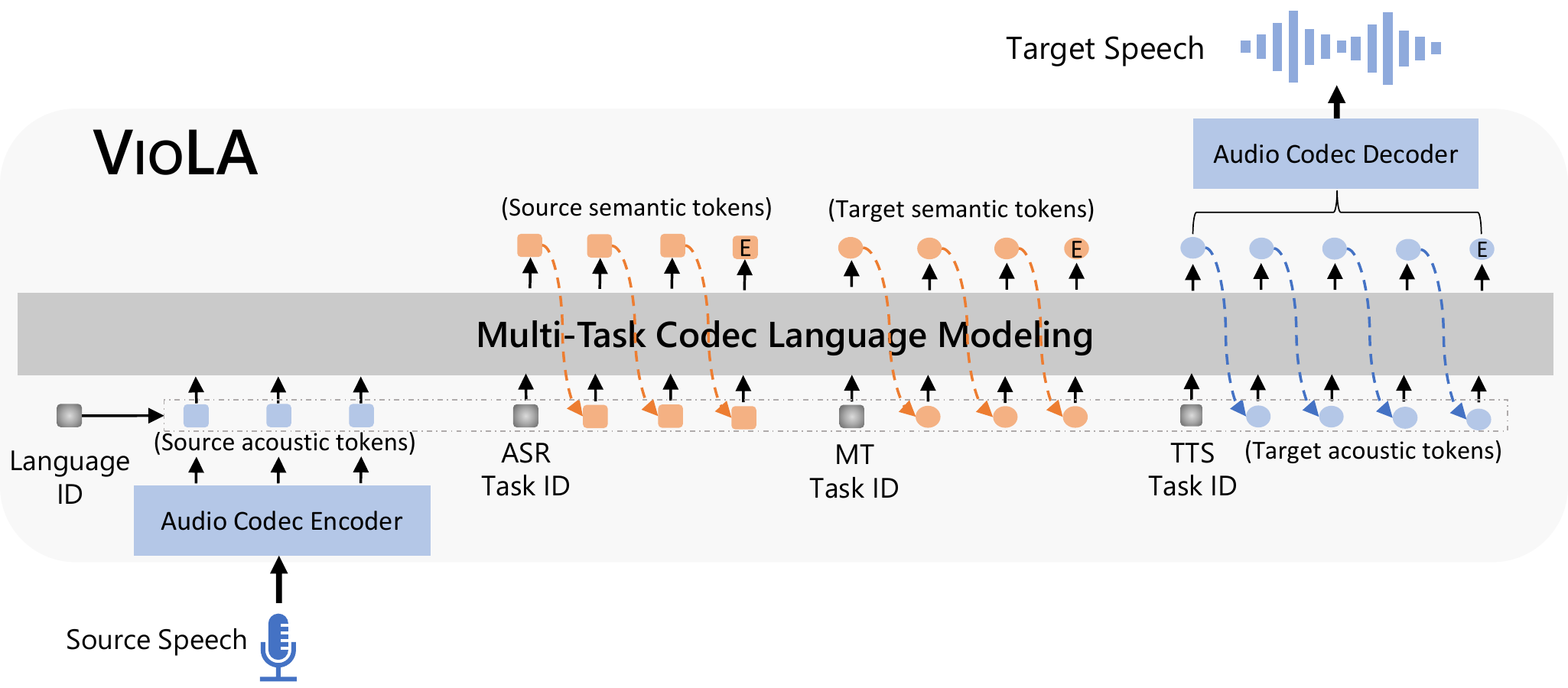}
  \caption{The overall framework of \our{}, which regards various speech processing tasks as a conditional codec language model task. 
  The model training is conducted on a multi-task learning framework with ASR, MT, and TTS tasks, and the model is capable of performing speech-to-text recognition and translation, text-to-text translation, text-to-speech synthesis, and speech-to-speech translation tasks.
  }  
  \label{fig_model_framework}
\end{figure*}



The success of generative models in NLP also extends to speech processing, namely speech synthesis tasks.
For instance, the VALL-E \cite{wang2023neural} model transforms continuous speech signal to discrete codec codes with EnCodec model \cite{encodec}, and trains a neural codec language model by scaling up TTS training data to 60K hours. VALL-E X \cite{zhang2023speak} extends this approach by introducing language ID to support cross-lingual TTS and speech-to-speech translation. The results show that with large training data, decoder-only models like VALL-E (X) can show in-context learning capacity and perform significantly better than baseline systems in zero-shot TTS tasks.
However, VALL-E (X) is only designed for the single speech synthesis task, and can not apply to other speech tasks.
This inspires us to ask one question: \textit{\textbf{Is one decoder-only generative model all you need for speech recognition, synthesis, and translation?}}

In this paper, we extend the application scenario of the decoder-only network from speech synthesis to all speech-processing tasks, including speech recognition, synthesis, and translation. 
Specifically, we present a multi-lingual multi-modal auto-regressive Transformer model \textbf{\our{}} to unify the speech-to-text, text-to-text, text-to-speech, and speech-to-speech tasks into a conditional codec language model task, as illustrated in Figure \ref{fig_model_framework}. By converting continuous speech signals to discrete codec codes via an offline neural codec model EnCodec, we can treat speech representation as textual tokens and employ a decoder-only model to optimize multi-modality tasks effectively.
Moreover, we leverage task IDs (TID) and language IDs (LID) to enhance the capability of distinguishing different languages and tasks.
We hope the proposed method can promote the era of codec-based speech processing with generative models.

We trained \our{} model under a multi-task learning framework, including automatic speech recognition (ASR), machine translation (MT), and text-to-speech synthesis (TTS) tasks. Massive evaluation on speech-to-text recognition and translation tasks, tex-to-text synthesis tasks, tex-to-speech translation tasks, and speech-to-speech translation tasks demonstrates that our proposed \our{} can be effectively applied to these single-modal and cross-modal tasks. 
Besides, the proposed model is able to maintain the in-context learning ability for speech synthesis tasks.
To the best of our knowledge, it is the first work to quantitatively explore codec-based auto-regressive Transformer decoder models for various speech processing tasks.

\section{Related Work}

Our work is built on an auto-regressive Transformer network for speech recognition, translation, and synthesis tasks.
In previous speech processing, different models with varying architectures were used for different tasks. The primary end-to-end model structures used for ASR \cite{E2EOverview, prabhavalkar2023end} are connectionist temporal classification (CTC) \cite{graves2006connectionist}, attention-based encoder-decoder network (AED) \cite{chiu2018state}, and transducer networks with RNN or Transformer \cite{chen2021developing}. 
For speech synthesis tasks, AED methods such as Tacotron \cite{wang2017tacotron}, Tacotron2 \cite{shen2018natural}, and TransformerTTS \cite{li2019neural}, and duration-based methods, including Fastspeech \cite{ren2019fastspeech}, Fastspeech2 \cite{ren2020fastspeech}, and RobuTrans \cite{li2020robutrans} are widely used.
At the same time, encoder-only models \cite{devlin2018bert}, decoder-only models \cite{radford2018improving}, and encoder-decoder models \cite{sutskever2014sequence,lewis2019bart} have also been extensively explored, and applied to various natural language processing tasks, such as machine translation, text summarization, question answer, and so on.

In terms of building one model for all speech tasks, HuBERT \cite{hsu2021hubert} and WavLM \cite{chen2022wavlm} learn a universal speech representation model using unsupervised masked unit prediction methods, like BERT \cite{devlin2018bert}. After pre-training, the pre-trained model, regarded as a speech representation extractor, and added specifical-task modal parameters, are fine-tuned using supervised data of downstream tasks.
Inspired by T5 \cite{raffel2020exploring}, SpeechT5 \cite{ao2021speecht5} adopts a universal encoder-decoder based sequence-to-sequence model for all spoken language tasks, e.g., speech recognition, speech translation, speech synthesis, etc.
To process different modalities, SpeechNet \cite{chen2021speechnet} uses multiple modal-specifical encoders and decoders for speech and text modalities.
Whisper \cite{radford2022robust} also employs an encoder-decoder framework trained with large-scale supervised data, but focuses on speech recognition and translation tasks.

Different from the above work, our proposed \our{} model tries to employ a Transformer decoder-only model for all spoken language tasks. Hence our work is also related to text-based and speech-based GPT models.
Generative pre-training (GPT) \cite{radford2018improving} methods have demonstrated the great potential capability for few-shot and zero-shot problems in ChatGPT \cite{chatgpt} and GPT-4 \cite{gpt4}. Moreover, researchers also apply the decoder-only model to text-to-speech synthesis tasks, namely VALL-E \cite{wang2023neural}, and VALL-E X \cite{zhang2023speak}.
They transform speech signals into discrete tokens, input cascaded discrete text and speech sequences to the Transformer decoder network, and show strong zero-shot transfer ability.
In parallel with our work, SpeechGPT \cite{zhang2023speechgpt} was recently proposed to perceive and generate multi-model content. However, it performs speech
discretization with hidden units from HuBERT, which contains less acoustic information than codec codes, and only lists some audio examples without quantitative results for cross-modal instruction following and speech dialogue tasks.
Our \our{} is based on VALL-E (X) and extends it to more speech tasks using the decoder-only Transformer model. It is also the first to quantitatively explore codec-based auto-regressive models for various spoken language tasks.

\section{Multi-Task Codec Language Model}

\subsection{Background}

Our proposed \our{} is built upon the neural codec language models VALL-E and VALL-E X, which treat text-to-speech synthesis as a language model task, like GPT, and employ acoustic tokens (audio codec codes) as an intermediate representation of original speech. VALL-E (X) generates the codec sequences (8 codes for each frame), based on which a codec decoder generates the final waveforms.  VALL-E (X) contains two key components, the auto-regressive codec language model and the non-autoregressive codec language model. The former is responsible for predicting the acoustic tokens of the first codec code for each frame based on the semantic tokens (phoneme sequences) in an auto-regressive manner, and the latter is used to generate the other 7-layer codes according to the sequence of the first-layer codes in parallel with the layer-level iterative generation method. 
In this work, we aim at extending VALL-E from the single TTS task to more speech tasks, e.g., speech recognition and speech-to-text translation, and focus on boosting the capability of the auto-regressive codec language model.

\begin{figure*}[h]
    \centering
    \includegraphics[width=15cm]{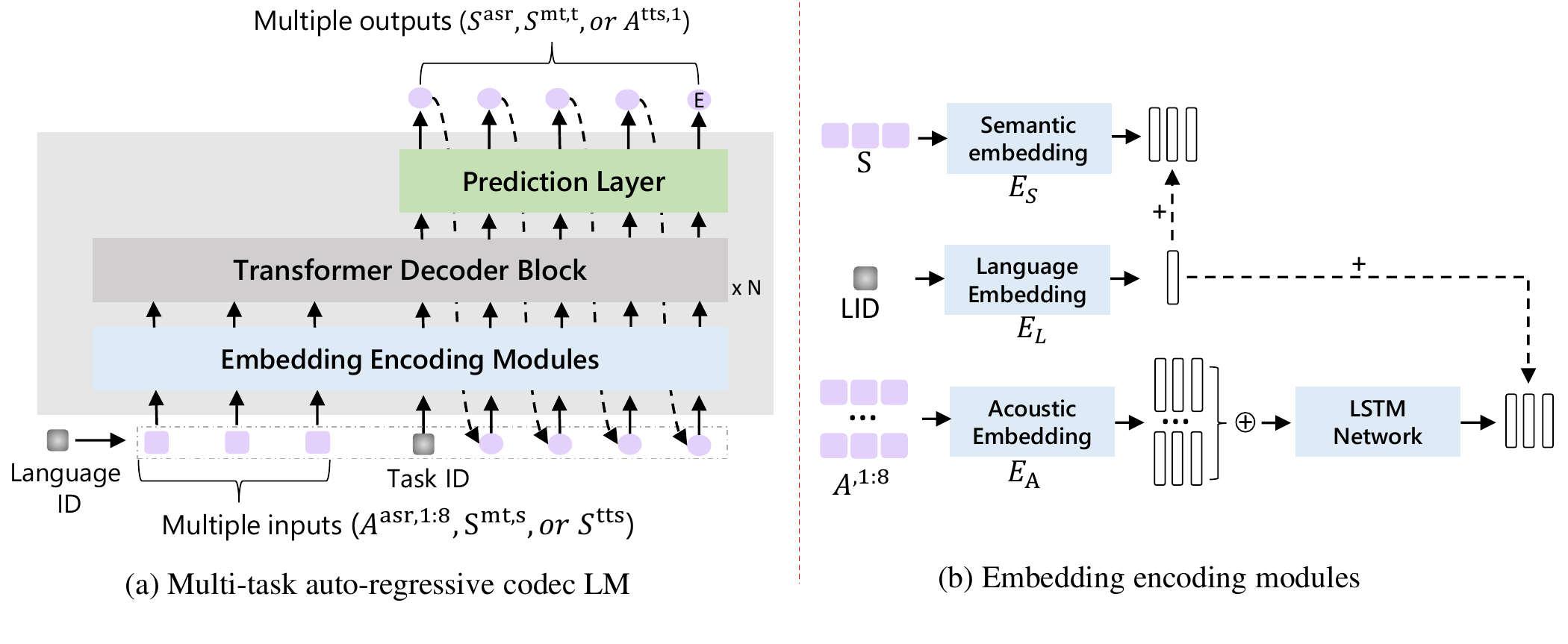}
    \vspace{-0.5cm}
    \caption{Left: Multi-task auto-regressive codec LM, consisting of embedding encoder modules, Transformer decoder block, and prediction layer. Right: Embedding encoding modules for semantic and acoustic tokens.}
    \label{AR_embedding}
\end{figure*}

\subsection{Problem Formulation}
\our{} regards all speech tasks as conditional
codec language modeling.
The core of our approach is to convert all the speech data into sequences of tokens (codec codes) as discrete text, and the speech-to-text, text-to-text, and text-to-speech tasks turn to token-based sequence conversion tasks with the decoder-only model.

Give the dataset $\mathcal{D} = \{\mathbf{x}^t_i , \mathbf{y}^t_i \}$, where $ t \in \{ \rm{asr, mt, tts} \}$ denotes different tasks, $i$ is the index of training samples, $\mathbf{x}$ and $\mathbf{y}$ denote audio sample and its transcription for ASR task, source text and target text for MT tasks, text transcription and audio sample for TTS task respectively. 
The speech data and text data in dataset $\mathcal{D} = \{\mathbf{x}^t_i , \mathbf{y}^t_i \}$ is first converted to codec codes and phonemes with an offline audio codec model (EnCodec) and grapheme to phoneme (G2P) conversion tool. 

More specifically, the speech side of the ASR and TTS training data, $\mathbf{x}^{\rm{asr}}$ and $\mathbf{y}^{\rm{tts}}$, are converted into discrete acoustic tokens, $\text{EnCodec}(\mathbf{x}^{\rm{asr}}) = \mathbf{A}^{{\rm{asr}}, 1:8} \in \mathbb{R}^{T_A \times 8}$,  $\text{EnCodec}(\mathbf{y}^{\rm{tts}}) = \mathbf{A}^{{\rm{tts}}, 1:8} \in \mathbb{R}^{T_A \times 8}$, where $\mathbf{A}$ is the 8-layer acoustic token matrix, and $T_A$ is the downsampled utterance length.
After quantization, the EnCodec decoder is capable of reconstructing the waveform from the acoustic tokens, denoted as $\text{Decodec}(\mathbf{A}) \approx \hat{\mathbf{y}}$.
For the text side of the training data, a G2P tool is used to convert $\mathbf{y}^{\rm{asr}}$, $\mathbf{x}^{\rm{mt}}$, $\mathbf{y}^{\rm{mt}}$, and $\mathbf{x}^{\rm{tts}}$ into their phoneme sequences $\mathbf{S}^{\rm{asr}}$, $\mathbf{S}^{\rm{mt, s}}$, $\mathbf{S}^{\rm{mt, t}}$, and $\mathbf{S}^{\rm{tts}} \in \mathbb{R}^{T_S}$, respectively, where $T_S$ is the length of phoneme sequences, and $\text{s}$ and $\text{t}$ in the upper right corner of $S$ represent source and target languages, respectively.

In this work, our goal is to leverage a conditional codec language model to model various spoken language tasks, and the model is optimized by maximizing $ p (\mathbf{y} |\mathbf{x}, \theta) = p ( \mathbf{S}^{\rm{asr}}|\mathbf{A}^{\rm{asr}}, \theta) + p  (\mathbf{S}^{\rm{mt, t}}|\mathbf{S}^{\rm{mt, s}}, \theta) + p ( \mathbf{A}^{\rm{tts}}|\mathbf{S}^{\rm{tts}}, \theta)$ like generative pre-training methods in GPT, with a multi-task learning framework.  
The trained \our{} is excepted to have the capability to conduct all the ASR, MT, TTS, cascaded speech-to-text translation, and cascaded speech-to-speech translation tasks.

\subsection{Model Framework}

\subsubsection{Multi-Task Auto-Regressive Codec LM}

Following VALL-E (X), an auto-regressive codec Transformer language model is leveraged as the core architecture of \our{},  shown in the left of Figure \ref{AR_embedding}.
It comprises an embedding encoding module, a Transformer decoder block, and a prediction layer.
Furthermore, we leverage language IDs to distinguish different languages and task IDs to accomplish different tasks, including speech recognition, machine translation, and speech synthesis.

The proposed auto-regressive codec LM is optimized by multiple tasks, using ASR corpus ($\mathbf{A}^{{\rm{asr}}, 1:8}$, $\mathbf{S}^{\rm{asr}}$), MT corpus ($\mathbf{S}^{\rm{mt, s}}$, $\mathbf{S}^{\rm{mt, t}}$), and TTS corpus ($\mathbf{S}^{\rm{tts}}$, $\mathbf{A}^{\rm{tts, 1}}$), as multiple inputs and outputs of Figure \ref{AR_embedding} (a). The multiple inputs, including semantic and acoustic tokens, are first processed using an embedding encoding module, which will be introduced in the next subsection (\ref{pre_post_net}) in detail.
After that, the standard Transformer decoder, consisting of the crucial multi-head attention network and feed-forward network, is employed to model the dependency relationship of semantic tokens and acoustic tokens. 
Note that all tokens in multiple inputs can attend over all positions in the input tokens, and each token in multiple outputs can attend to previous multiple outputs and all input tokens.
Finally, we use a prediction layer to convert hidden states into the discrete index of vocabulary.

\subsubsection{Pre-Net and Pos-Net}
\label{pre_post_net}

\paragraph{Embedding Encoding Modules} 

To better represent original acoustic tokens and semantic tokens, we design embedding encoding modules to encode the diverse discrete tokens, as illustrated in the right of Figure \ref{AR_embedding}.
For the semantic tokens $\mathbf{S} \in \mathbb{R}^{T_\text{S}}$, including $\mathbf{S}^{\rm{asr}}$, $\mathbf{S}^{\rm{mt, s}}$, $\mathbf{S}^{\rm{mt, t}}$, and $\mathbf{S}^{\rm{tts}}$, we use semantic embedding matrix $\text{E}_S$ to obtain the regular semantic embedding. In the same way, the language ID $L\in \mathbb{R}^{1}$ is embedded with $\text{E}_L$ and incorporated into the semantic embedding as,
\begin{equation}
\label{Xs}
\bm{X_\text{S}} = \text{E}_S(\bm{S}) + \text{E}_L(L)
\end{equation}
where $\bm{X_\text{S}} \in \mathbb{R}^{T_\text{S}\times D}$, e.g., $\bm{X_\text{S}}^{\rm{asr}}$, $\bm{X_\text{S}}^{\rm{mt,s}}$, $\bm{X_\text{S}}^{\rm{mt, t}}$, and $\bm{X_\text{S}}^{\rm{tts}}$, are the semantic feature through embedding encoding modules, and $D$ is the dimension of hidden states.

For the acoustic tokens $\mathbf{A}^{{\rm{asr}}, 1:8}$ and $\mathbf{A}^{\rm{tts, 1}}$, we also first get the acoustic embedding through an acoustic embedding matrix $\text{E}_{A, i}$, where $i$ denotes the $i$-th layer of acoustic tokens.
Moreover, we average all-layer acoustic embedding to a mixed embedding feature to represent multi-layer acoustic embedding.
Inspired by EnCodec decoder, the mixed embedding feature of $\mathbf{A}^{{\rm{asr}}, 1:8}$ and single embedding feature of $\mathbf{A}^{\rm{tts, 1}}$  are finally fed into a unidirectional LSTM to obtain the acoustic features $\bm{X_\text{A}} \in \mathbb{R}^{T_\text{A} \times D}$, as follows,
\begin{align}
\label{Xa}
\bm{X_\text{A}^{\text{asr}}} &= \text{LSTM}\left( \sum^8_{i=1}{\text{E}_{A,i}(\bm{A}^\text{asr,i})}/8\right) + \text{E}_L(L) \\
\label{equa_lstm}
\bm{X_\text{A}^{\text{tts,1}}} &= \text{LSTM}\left( \text{E}_{A,1}(\bm{A}^\text{tts,1})\right) + \text{E}_L(L)
\end{align}

\paragraph{Non-Autoregressive Codec LM} 
For speech synthesis tasks, the above auto-regressive code LM only estimates the first-layer acoustic tokens. Following VALL-E (X), we adopt another non-auto-regressive codec Transformer language model to generate the all-layer acoustic tokens.
For simplicity, we refer readers to read the non-auto-regressive codec LM part in VALL-E (X).
Different from the original non-autoregressive codec LM, the LSTM module is also introduced for encoding acoustic tokens in our non-autoregressive codec language model, as Equation \ref{equa_lstm}.

\subsection{Training Objectives}
The multi-task auto-regressive codec language model is optimized by performing speech-to-text recognition, text-to-text translation, and text-to-speech synthesis tasks, as shown in Figure~\ref{AR_embedding}. Particularly, we integrate additional task IDs (e.g., $b_{\text{asr}}$, $b_{\text{mt}}$, $b_{\text{tts}}$) into the proposed model to accomplish different tasks.

\paragraph{Speech-to-Text Recognition Task} aims to predict the phoneme sequences $\bm{S}^\text{asr}$ according to the acoustic tokens $\bm{A}^\text{asr,1:8}$ from the original speech. Here, eight layers of the acoustic tokens are employed as the input to introduce more acoustic information. The training is implemented as teacher forcing and auto-regressive strategies with an ASR task ID $b_{\text{asr}}$, as
\begin{equation}
\begin{split}
\mathcal{L}_{asr} &= p\left(\bm{S}^\text{asr}|\bm{A}^\text{asr,1:8}, b_{\text{asr}};\theta\right) \\
&=\prod_{t=0}^{T_\text{S}}p\left(S_{t}^\text{asr}|\bm{A}^\text{asr,1:8},b_{\text{asr}},\bm{S}_{<t}^\text{asr},\theta\right) 
\end{split}
\end{equation}
where the acoustic tokens, task ID, and semantic tokens are concatenated and then fed into the language model.

\paragraph{Text-to-Text Translation Task}
is to translate the source phoneme sequences $\bm{S}^\text{mt,s}$ into the target language phonemes $\bm{S}^\text{mt,t}$. Similar to speech recognition, MT task is optimized by maximizing the following probability,

\begin{equation}
\begin{split}
\mathcal{L}_{mt} &= p\left(\bm{S}^\text{mt,t}|\bm{S}^\text{mt, s}, b_{\text{mt}};\theta \right) \\
&= \prod_{t=0}^{T_\text{S}}p\left(S_{t}^\text{mt, t}|\bm{S}^\text{mt,s},b_{\text{mt}},\bm{S}_{<t}^\text{mt,t},\theta\right) \\
\end{split}
\end{equation}

\paragraph{Text-to-Speech Synthesis Task} focuses on generating the first-layer quantized acoustic tokens $\bm{A}^\text{tts,1}$ depending on their corresponding transcription $\bm{S}^\text{tts}$ with auto-regressive manner, as follows,

\begin{equation}
\begin{split}
\mathcal{L}_{tts} &= 
p\left(\bm{A}^\text{tts,1}|\bm{S}^\text{tts}, b_{\text{tts}};\theta \right)  \\
&=\prod_{t=0}^{T_\text{A}}p\left(A_{n}^\text{tts,1}|\bm{S}^\text{tts},b_{\text{tts}},\bm{A}_{<t}^\text{tts,1},\theta \right) 
\end{split}
\end{equation}

\paragraph{Multi-Task Learning}
We optimize the auto-regressive codec LM with a multi-task learning framework, namely,
\begin{equation}
\mathcal{L} = \mathcal{L}_{asr} + \mathcal{L}_{mt} + \mathcal{L}_{tts}
\end{equation}
where these three tasks are simultaneously trained.

\subsection{Inference Methods}
\label{inference_method}
After training, our model can be applied to various tasks, including ASR, MT, TTS, speech-to-text translation (S2TT), and speech-to-speech translation (S2ST) tasks.
For the last two tasks, we adopt a cascaded pipeline, consisting of the first three base tasks. 
We leave end-to-end speech-to-text translation and speech-to-speech translation for future work.
It is noted that we use sampling methods for speech synthesis tasks, and beam search for other tasks.
For speech synthesis tasks, we can use speech utterances with the same language or different languages as speech prompts, like VALL-E and VALL-E X. We perform five synthesis inferences for one text, and the utterance with the highest score is selected according to speaker similarity (SS) score, aka Strategy I, or the combination of speaker similarity and word error rate (WER) scores, aka Strategy II\footnote{Strategy I chooses the utterance with the highest speaker similarity. Strategy II chooses the utterance with the highest score of SS minus WER (SS and WER are the normalized scores of the five results), and the calculation of SS and WER will be introduced in Section \ref{metrics}.}.

\section{Experiments}

We evaluate our proposed \our{} on tasks of ASR, MT, S2TT, zero-shot TTS, and zero-shot S2ST.

\subsection{Data}

Our proposed \our{} is trained on two speech recognition datasets, which can also be used to train TTS tasks as VALL-E (X), and two machine translation datasets. The Chinese ASR data are from WenetSpeech \cite{wenetspeech}, a total of 10,000 hours of multi-domain labeled speech. The English ASR data are from LibriLight \cite{librilight}, a total of 60,000 hours of unlabeled audiobook speech, where the transcripts are generated by an ASR model\footnote{\url{https://github.com/kaldi-asr/kaldi/tree/master/egs/librispeech}} trained on the LibriSpeech \cite{librispeech}. AI Challenger\footnote{\url{https://challenger.ai/competition/translation}} and WMT2020\footnote{\url{https://www.statmt.org/wmt20/translation-task.html}} datasets are employed for MT training, containing 63M English-Chinese sentence pairs in conversion, drama, and news domains. 

We evaluate our proposed method on WenetSpeech, EMIME \cite{wester2010emime}, LibriSpeech, and the WMT2020 datasets. EMIME contains 25 pairs of bilingual sentences recorded by seven female and seven male native Chinese speakers with two microphones. We evaluate speech-to-phoneme ASR on the development set of WenetSpeech. The MT performance is evaluated on the test set of WMT2020. The Chinese-to-English S2TT (ZH$\Rightarrow$EN S2TT), zero-shot English TTS prompted by Chinese speech (ZH$\mapsto$EN TTS), and zero-short Chinese-to-English S2ST (ZH$\Rightarrow$EN S2ST) are evaluated on the EMIME dataset. The zero-shot English TTS prompted by English speech (EN TTS) is evaluated on the Librispeech dev-clean and test-clean sets.


\subsection{Data Pretreatment}
\label{data_pre}
The proposed \our{} is trained on two types of discrete tokens, semantic tokens (phoneme sequences) and acoustic tokens (codec codes). The transcription of ASR data and the bilingual text of MT are converted into phonemes via the lexicon provided by ASR datasets and the International Phonetic Alphabet (IPA)-based unified phoneme set called BigCiDian\footnote{\url{https://github.com/speechio/BigCiDian}}. We also use the Kaldi force-alignment tool\footnote{\url{https://github.com/kaldi-asr/kaldi/tree/master}} to generate the alignment information for clipping utterances during training. The speech data are quantized into acoustic tokens using the EnCodec\footnote{\url{https://github.com/facebookresearch/encodec}} with 6 kbps bandwidth in our method, resulting in 8 tokens for each frame. 
We also pre-process speech into the 80-dimensional feature with a frameshift of 10ms for baselines.

\subsection{Model Architecture}
The auto-regressive codec language model employs a Transformer decoder with an attention dimension of 1024 and the FFN dimension of 4096. Sinuous position embedding is separately computed for multiple input and output sequences in Figure \ref{AR_embedding} (a). 
In embedding encoding modules, we use layer-specific 1024-dimensional embeddings for each layer of acoustic tokens, and one 1024-dimensional embedding for semantic tokens. Besides, 3-layer unidirectional LSTM is employed for this pre-net. 
Since \our{} is required to support more tasks, which may need larger model capability with more parameters, we trained 12-layer and 18-layer decoder-only auto-regressive models for comparison, called \textbf{\our{} (12L)} and \textbf{\our{} (18L)}, with 178M and 250M parameters, respectively.

We built two strong baselines, including encoder-decoder (\textbf{AED}) models and decoder-only language model (\textbf{LM}) for comparison.
AED models adopt 6 Transformer encoder and decoder layers, with additional cross-attention modules, for MT and Fbank-based ASR tasks. 
Two-layer convolution with 4-times down-sampling of time dimension is employed to process the Fbank before Transformer. 
We trained the Fbank-based ASR models without SpecAugment \cite{park2019specaugment} to compare the performance of different input features for ASR fairly. 
The configuration of LMs on single tasks is the same with 12-layer \our{}.

\subsection{Training Details}
Our multi-task auto-regressive codec language model is trained on ASR, MT, EN TTS, and ZH TTS total of 4 tasks simultaneously for 800K steps on 32 V100 GPUs with a batch size of 25 seconds per GPU (the losses of 4 tasks are accumulated for one step). 
We re-segment the training data to an average utterance duration of 12 seconds for effective training, and the maximum sentence length is 20 seconds.
Each batch loads MT data with the same number of semantic tokens as the number of 25s acoustic tokens. The maximum learning rate is $5\times 10^{-4}$ with warm-up step of 80K.  
We follow the configuration of VALL-E X to train our non-autoregressive language model, which integrates additional embedding encoding modules as introduced in Section \ref{pre_post_net}.

\subsection{Evaluation Metrics}
\label{metrics}
We employ phoneme error rate (PER) and BLEU to evaluate the performance of models on speech recognition tasks and three translation tasks (MT, S2TT, and S2ST). 
Instead of human evaluation, we use the automatic evaluation metrics, including the word error rate (WER), speaker similarity (SS), and speech naturalness (SN) for EN TTS to evaluate the generated speech for simplicity and convenience.
For synthesized speech in TTS and S2ST tasks, we first utilize a HuBERT-Large ASR model\footnote{\url{https://github.com/facebookresearch/fairseq/tree/main/examples/hubert}} finetuned on LibriSpeech to transcribe it into text, then calculate the above WER and BLEU scores. 
Given generated and prompt speech utterances, the SS is measured by an automatic speaker verification (ASV) model\footnote{\url{https://github.com/microsoft/UniSpeech/tree/main/downstreams/speaker_verification}}, ranging from -1 to 1. The larger the SS, the more similar the speakers of the two utterances are. 
SN score of generated speech is measured by the open-source NISQA\footnote{\url{https://github.com/gabrielmittag/NISQA}} \cite{mittag2021deep}.

\subsection{Main Results}

\paragraph{Speech Recognition Evaluation}
\label{asr_exp}
We first evaluate the speech recognition performance of different models on the development set of WenetSpeech. 
The results are summarized in Table \ref{asr_table}.
The Fbank-based AED achieves the lowest PER. Fbank-based decoder-only LM causes a slight increase in the PER, which proves that the LM can still achieve comparable performance on the speech recognition task. 
As can be seen from the results of \our{} (12L),  integrating multiple tasks into a single model slightly impairs the performance of speech recognition tasks. However, \our{} (18 L), with comparable parameters to the AED, achieves acceptable results for codec-based speech recognition with a PER of 11.36.

\begin{table}[htp]
\centering
\small
\begin{tabular}{lcccc}
\toprule
Method & Input  & Param.(M) & PER $\downarrow$\\
\midrule
AED (enc-dec) & \multirow{2}{*}{Fbank} & 246.3 & \textbf{9.47}   \\
LM (decoder) &   & \textbf{150.8} & 9.61   \\
\midrule
LM (decoder) & \multirow{3}{*}{Codec} & \textbf{177.5} & 11.71   \\
\our{} (12L) &  & 178.5 & 12.97   \\
\our{} (18L) & & 250.6 & \textbf{11.36} \\
\bottomrule
\end{tabular}
\caption{Comparison of different models on speech recognition task.}
\label{asr_table}
\end{table}



\paragraph{Machine Translation Evaluation}

We verify the performance of different models on the machine translation task based on the test set of WMT2020 after processing in subsection~\ref{data_pre}, and the results are shown in Table~\ref{mt_table}.
LM (decoder-only model) achieves comparable results to the AED (encoder-decoder model) on the machine translation task.
\our{} (18L) can obtain the improvement of +2.53 BLEU scores than \our{} (12L).
Furthermore, \our{} (18L) achieves the highest BLEU score (56.97) with a comparable number of parameters as AED models (250.6M vs. 242.5M).

\begin{table}[htp]
\centering
\small
\begin{tabular}{lccc}
\toprule
Method &  Param.(M) & BLEU $\uparrow$ \\
\midrule
AED (enc-dec) &  242.5 & 56.83   \\
LM (decoder)    & \textbf{145.4} & 56.81   \\
\our{} (12L)   & 178.5 & 54.44   \\
\our{} (18L) & 250.6 & \textbf{56.97} \\
\bottomrule
\end{tabular}
\caption{Comparison of different models on machine translation task.}
\label{mt_table}
\end{table}

\paragraph{Speech-to-Text Translation Evaluation}
We perform ZH$\Rightarrow$EN S2TT tasks on the EMIME dataset by cascading ASR and MT tasks of \our{}.
As listed in Table~\ref{s2t_table}, the BLEU score of the codec-based cascaded LM is only 0.28 lower than the BLEU of 55.98 achieved by the Fbank-based cascaded AED models. We speculate that comparable MT performance can compensate for the shortcomings of ASR performance for our model. 
Moreover, \our{} (18L) achieves a comparable BLEU score on the S2TT task (55.85 vs. 55.98) with a 48.7\% reduction in parameters compared to the Fbank-based cascaded AED models.

\begin{table}[htp]
\setlength\tabcolsep{5.0pt}
\centering
\small
\begin{tabular}{lccc}
\toprule
Method  & Input & \makecell{Total Param.(M)}  & BLEU $\uparrow$\\
\midrule
$\text{AED}_{\text{ASR}}$ $\veryshortarrow$ $\text{AED}_{\text{MT}}$ & \multirow{1}{*}{\makecell{Fbank}} & 488.8 &  \textbf{55.98}  \\
\midrule
$\text{LM}_{\text{ASR}}$ $\veryshortarrow$ $\text{LM}_{\text{MT}}$ & \multirow{3}{*}{\makecell{Codec}} & 322.9  & 55.70  \\
\our{} (12L) &  & \textbf{178.5} & 53.16 \\
\our{} (18L)  & & 250.6 & \textbf{55.85} \\
\bottomrule
\end{tabular}
\caption{Comparison of different models on speech-to-text translation task.}
\label{s2t_table}
\end{table}


\begin{table*}[htp]
\centering
\small
\begin{tabular}{lccccccccc}
\toprule
  \multirow{2}*{Method} & \multicolumn{3}{c}{Strategy I} & \multicolumn{3}{c}{Strategy II} & \multicolumn{3}{c}{AVG}  \\
  \cmidrule(rl){2-4}\cmidrule(rl){5-7}\cmidrule(rl){8-10}
  & SS & WER & SN & SS & WER & SN & SS $\uparrow$ & WER $\downarrow$ & SN $\uparrow$\\
\midrule
Ground Truth Audio & - & - & - & - & - & - & 0.67 & 1.98 & 3.81 \\
\midrule
VALL-E~X &  0.53 & 5.81 & 3.20 & 0.49 & 3.38 & 3.20 & 0.51 & 4.60 & 3.20  \\
\our{} (12L) & 0.52 & 6.13 & 3.20 & 0.47 & 3.75 & 3.19 & 0.50 & 4.94 & 3.20 \\
\our{} (18L)   & \textbf{0.54} & \textbf{4.97} & \textbf{3.22} & \textbf{0.50} & \textbf{2.89} & \textbf{3.21} & \textbf{0.52} & \textbf{3.93} & \textbf{3.22}  \\
\bottomrule
\end{tabular}
\caption{Comparison of different models on zero-shot text-to-speech task. SS means speaker similarity, SN means speech naturalness, AVG means the average scores of Strategy I and Strategy II.}
\label{tts_table}
\end{table*}

\paragraph{Zero-Shot Text-to-Speech Synthesis Evaluation}
\label{ttssection}
As mentioned in subsection~\ref{inference_method}, we synthesize the English speech of corresponding text  prompted by an English speech utterance on selected samples of dev-clean and test-clean sets\footnote{To keep the consistency between training and inference, we select 1802 samples of 4 to 10 seconds with the same speaker's speech prompt of 3 to 5 seconds from the LibriSpeech dev-clean and test-clean sets to evaluate different models on the zero-short EN TTS task.}.
As shown in Table~\ref{tts_table}, \our{} (18L) significantly outperforms \our{} (12L) in all ASR, MT, and TTS tasks, which suggests that LM with more tasks and larger data may require the larger model to train.
Compared to the VALL-E X, the speaker similarity of \our{} (18L) is relatively improved by 2.0\%, the WER is relatively decreased by 14.6\%, and the speech naturalness is improved by 0.02.
Our model also shows powerful in-context learning capabilities in keeping the speaker's similarity of prompt speech.

\paragraph{Zero-Shot Cross-Lingual Text-to-Speech Synthesis Evaluation}
Different from EN TTS, zero-shot cross-lingual ZH$\mapsto$EN TTS is to synthesize English speech by using both the English text and the Chinese speech as prompts, like VALL-E X.
Experimental results are summarized in Table~\ref{s2st_table}.
\our{} (18L) improves the average SS by 0.01 and BLEU by 2.20 compared to the VALL-E X, and generates more stable codecs for natural English speech with a 3.35 naturalness score.


\begin{table*}[htp]
\setlength\tabcolsep{5.0pt}
\centering
\small
\begin{tabular}{lcccccccccc}
\toprule
  \multirow{2}*{Method} & \multirow{2}*{Input}  &  \multicolumn{3}{c}{Strategy I} & \multicolumn{3}{c}{Strategy II} & \multicolumn{3}{c}{AVG} \\
  \cmidrule(rl){3-5}\cmidrule(rl){6-8}\cmidrule(rl){9-11}
   &  & SS & BLEU & SN & SS & BLEU & SN & SS $\uparrow$& BLEU $\uparrow$& SN $\uparrow$\\
\midrule
Ground Truth Audio & - & - & - & - & - & - & - & 0.58 & 93.31 & 3.82 \\
\midrule
\textit{Zero-shot cross-lingual text-to-speech} \\
\hline
Target text $\veryshortarrow $ VALL-E~X & \multirow{3}*{\makecell{Text}}  & 0.49 & 69.37 & 3.32 & 0.44 & 83.99 & 3.35 & 0.47 & 76.68 & 3.34 \\
Target text $\veryshortarrow $ \our{} (12L) & & 0.49 & 65.22 & 3.32 & 0.43 & 82.42 & 3.31 & 0.46 & 73.82 & 3.32 \\
Target text $\veryshortarrow $ \our{} (18L) &  & \textbf{0.50} & \textbf{72.55} & \textbf{3.33} & \textbf{0.45} & \textbf{85.21} & \textbf{3.36} & \textbf{0.48} & \textbf{78.88} & \textbf{3.35}\\
\midrule
\textit{Zero-shot speech-to-speech} \\
\hline
$\text{AED}_{\text{ASR}}$ $\veryshortarrow$ $\text{AED}_{\text{MT}}$  $\veryshortarrow$ 
VALL-E~X  &  \makecell{Fbank} &  0.50 & 42.00 & 3.52 & 0.48 & 49.30 & 3.53  & 0.49 & 45.65 & 3.53\\
\hdashline[2.5pt/5pt]
$\text{LM}_{\text{ASR}}$ $\veryshortarrow$ $\text{LM}_{\text{MT}}$ $\veryshortarrow$ VALL-E~X   & \multirow{3}*{\makecell{Codec}} & 0.50 & 41.05 & 3.51 & 0.48 & 48.74 & 3.52  & 0.49 & 44.90 & 3.52\\
\our{} (12L) & & 0.49 &  39.26 & 3.49 & 0.47 & 45.86 & 3.52 & 0.48 & 42.56 & 3.51\\
\our{} (18L) & & \textbf{0.51} & \textbf{43.96} & \textbf{3.54} & \textbf{0.49} & \textbf{51.57} & \textbf{3.56} & \textbf{0.50} & \textbf{47.77} & \textbf{3.55}\\
\bottomrule
\end{tabular}
\caption{Comparison on zero-shot cross-lingual text-to-speech and speech-to-speech translation tasks.}
\label{s2st_table}
\end{table*}

\paragraph{Zero-Shot Speech-to-Speech Translation Evaluation}
We conduct cascaded ASR, MT, and TTS tasks of \our{} to accomplish ZH$\Rightarrow$EN S2ST.
Compared with the codec-based cascaded LMs, the Fbank-based cascaded AED with VALL-E X achieves an average 0.76 improvement in the BLEU score with its advantage in speech recognition.
\our{} (18L) with 1.5 times parameters than \our{} (12L) balances three single tasks and achieves the best performance on the zero-short ZH$\Rightarrow$EN S2ST task with an average SS of 0.50, an average BLEU of 47.27, and an average speech naturalness of 3.55.

\subsection{Analysis}
In this section, we will analyze the effect of the layer number of acoustic tokens and LSTM networks in embedding encoder modules for speech recognition and synthesis, respectively.

\paragraph{Effect of Layer Number of Acoustic Tokens}
As introduced in Section \ref{pre_post_net}, we average the acoustic embeddings of the eight codecs for each frame, resulting in the 1024-dimensional frame-level input for Transformer. We first explore the effect of the number of acoustic codecs per frame for speech recognition, and the results are shown in Table~\ref{asr_ab_table}.
As we can see from the results, the increase in the layer number of acoustic tokens significantly improves speech recognition performance. Compared with one layer of acoustic tokens, eight layers of acoustic tokens provide more speech information, and the phoneme error rate is relatively reduced by 54.5\%, with only seven additional 1024-dimensional embeddings.

\begin{table}[htp]
\centering
\small
\begin{tabular}{lcccc}
\toprule
Method & Codec Layers & Param.(M) & PER \\
\midrule
\multirow{4}*{\makecell{LM}} & 0 & \textbf{147.5} &  28.21  \\
 & 0, 1 & 148.5 & 19.32  \\
 & 0, 1, 2 & 149.5 & 14.16 \\
 & 0, 1, ... , 7 & 153.5 & 12.83\\
 \midrule
 LM + LSTM$_{\text{AT}}$ & 0, 1, ... , 7 & 177.5 & \textbf{11.71} \\
\bottomrule
\end{tabular}
\caption{Comparison of LMs with different numbers of codec layers on ASR.}
\label{asr_ab_table}
\end{table}

\paragraph{Effect of LSTM Network}
Inspired by the decoder of EnCodec, \our{} employs an additional LSTM network to process the acoustic embedding. The results in Table~\ref{asr_ab_table} show that, given the 8-layer acoustic tokens, the introduction of LSTM for embedding encoding modules in the auto-regressive model (LSTM$_{\text{AT}}$) relatively reduces the PER by 8.7\% on ASR task.
We also investigate the influence of LSTM networks in TTS tasks for both auto-regressive and non-auto-regressive models.
The results of the zero-shot TTS ablation experiments are shown in Table~\ref{tts_ab_table}, which demonstrates that the integration of LSTM networks in both auto-regressive and non-autoregressive models improves the performance of the decoder-only model on speech synthesis tasks. The average WER of \our{} with the $\text{LSTM}_{\text{AT+NAT}}$ is relatively reduced by 12.5\% compared to the VALL-E X.

\begin{table}[htp]
\centering
\small
\begin{tabular}{lcccc}
\toprule
  \multirow{2}*{Method} & \multicolumn{2}{c}{Strategy I} & \multicolumn{2}{c}{Strategy II} \\
  \cmidrule(rl){2-3}\cmidrule(rl){4-5}
  & SS & WER & SS & WER \\
\midrule
LM (VALL-E~X) &  0.53 & 5.81 & 0.49 & 3.38  \\
\ \  + LSTM$_{\text{AT}}$ &  0.53 & 5.29 & \textbf{0.49} & 3.17  \\
\ \  + LSTM$_{\text{AT+NAT}}$ & \textbf{0.53} & \textbf{5.02} & 0.48 & \textbf{3.02} \\
\bottomrule
\end{tabular}
\caption{Comparison of LMs with the different acoustic embedding processes on TTS.}
\label{tts_ab_table}
\end{table}

\section{Conclusion}
In this paper, we propose \our{}, an auto-regressive codec language model for multi-modal tasks involving speech and text.
\our{} is the first work to quantitatively explore and apply a Transformer decoder-only model and codec-based methods to various spoken language tasks.
By discretizing speech signals to discrete acoustic tokens as textual semantic tokens, all spoken processing tasks can be formulated into one conditional language modeling task with the same optimization objective. Massive experiments on speech-to-text recognition and translation, machine translation, text-to-speech synthesis, and speech-to-speech translation demonstrate the effectiveness and superiority of the proposed model.

\section{Limitations}

Although the proposed conditional codec language model \our{} is successfully applied to various speech tasks, e.g., speech recognition, synthesis, and translation tasks, it still suffers from some limitations:
(1) It is optimized with supervised data and does not make full use of unsupervised data, such as large-scale text corpora and unlabeled speech.
(2) The model only shows in-context learning ability on speech synthesis tasks, rather than all speech processing tasks.
(3) Current \our{} only supports cascaded inference methods for speech-to-text translation tasks, and speech-to-speech translation tasks, without end-to-end capability.
 
\bibliography{anthology,custom}
\bibliographystyle{acl_natbib}




\end{document}